# A Review of Physics-based Machine Learning in Civil Engineering

Shashank Reddy Vadyala[1]; Sai Nethra Betgeri[1]; Dr. John C. Matthews[2]; Dr. Elizabeth Matthews[3]

1. Department of Computational Analysis and Modeling, Louisiana Tech University, Ruston, Louisiana, United States.
2. Director, TTC, Louisiana Tech University, Ruston, Louisiana, United States.
3. Assistant Professor, Civil Engineering, Louisiana Tech University, Ruston, Louisiana, United States.

**Abstract:**

The recent development of machine learning (ML) and Deep Learning (DL) increases the opportunities in all the sectors. ML is a significant tool that can be applied across many disciplines, but its direct application to civil engineering problems can be challenging. ML for civil engineering applications that are simulated in the lab often fail in real-world tests. This is usually attributed to a data mismatch between the data used to train and test the ML model and the data it encounters in the real world, a phenomenon known as data shift. However, a physics-based ML model integrates data, partial differential equations (PDEs), and mathematical models to solve data shift problems. Physics-based ML models are trained to solve supervised learning tasks while respecting any given laws of physics described by general nonlinear equations. Physics-based ML, which takes center stage across many science disciplines, plays an important role in fluid dynamics, quantum mechanics, computational resources, and data storage. This paper reviews the history of physics-based ML and its application in civil engineering.

Keywords: Physics-based machine learning, Machine Learning, Deep neural network, Civil engineering

# 1. Introduction

ML and DL, e.g., deep neural networks (DNNs), are becoming increasingly prevalent in the scientific process, replacing traditional statistical methods and mechanistic models in various commercial applications and fields, including education [1], natural science [2, 3] medical [4-6] engineering [7-9], and social science[10]. ML is also applied in civil engineering, where mechanistic models have traditionally dominated [11-14]. Despite its wide adoption, researchers and other end users often criticize ML methods as a "black box," meaning they are thought to take inputs and provide outputs but not yield physically interpretable information to the user[15]. As a result, some scientists have developed physics-based ML to reckon with widespread concern about the opacity of black-box models [16-19].

The civil engineering ML models are created directly from data by an algorithm; even researchers who design them cannot understand how variables are combined to make predictions. Even with a list of input variables, black-box predictive ML models can be such complex functions that no researchers can understand how the variables are connected to arrive at a final prediction. For example, ML models that fail to estimate structural damage are tied to processes that are not entirely understood have difficulty providing high data needs. Hence, their high data needs, difficulty providing physically consistent findings, and lack generalizability to out-of-sample scenarios [18]. Large, curated data sets with well-defined, precisely labeled categories are used to test ML and DL model[1]s. DL does well for these problems because it assumes a largely stable world. But in the real world, these categories are constantly evolving, specifically in civil engineering. Only after extensive testing on ML responses to various visual stimuli we can discover the problem.

Physics-based numerical simulations have become indispensable in civil engineering applications, such as seismic risk mitigation, irrigation management, structural design and analysis, and structural health monitoring. Civil engineers and scientists may now utilize sophisticated models for real-world applications, with ultra-realistic simulations involving millions of degrees of freedom, thanks to the advancement of high-performance computers. However, in the civil engineering sector, such simulations are too time-consuming to be incorporated fully into an iterative design process. They are often restricted to the final validation and certification stages, while most design processes rely on simpler models. Accelerating complex simulations is an important problem to address since it would make it easier to apply numerical tools throughout the design process. The development of numerical methods for rapid simulations would also enable novel model applications such as improving construction productivity, which has yet to be fully utilized due to model complexity. Uncertainty quantification is another critical example of analysis that might be feasible if simulation costs were lowered substantially. Indeed, the physical system environment, which is generally unknown, affects the values of interest monitored in numerical simulations. In some situations, these uncertainties significantly impact simulation results, necessitating estimating probability distributions for the quantities of interest to assure the product's dependability. Neither an ML-only nor a scientific knowledge-only method can be considered sufficient for complicated scientific and technical applications. Researchers are beginning to investigate the continuum between mechanistic and ML models, synergizing scientific knowledge and data.

There have been several reviews on ML civil engineering. However, limited studies have been conducted on physics-based ML and synthesizing a road map for guiding subsequent research to advance the proper use of physics-based ML in civil engineering applications. Furthermore, there are few works focused on the fundamental physics-based ML models in civil engineering. This study investigates a more profound connection of ML methods with physics models. Even though the notion of combining scientific principles with ML models has only recently gained traction[18], there has already been a significant amount of research done on the subject. Researchers focus on physics models, ML models, and application scenarios to solve their problems in civil engineering. This study aims to bring these exciting developments to the ML community and make them aware of the progress completed and the gaps and opportunities for advancing research in this promising direction.

**Basics of Neural Network and Physics-Based Machine Learning**

Neural Networks (NNs) is a ML approach for expressing a form's input-output relationship, shown in Equation (1)

$$(1)$$
$$Y = Y^{NN} = W^T \emptyset_h(B^T \bar{x}) + \eta$$

where $\bar{x} = [x; 1]$, y is the target (output) variable, while x is the input variable, $Y^{NN}$ is the predicted output variable obtained from NNs. The input variable's activation function is $\emptyset_h$, the transition weight matrix is U, the output weight matrix is W, and W is an unknown error owing to measurement or modeling mistakes are W. Within the weight matrices, the bias terms are defined by supplementing the input variable x with a unit value in the present notations. In Equation (1), the target variable is a linear combination of certain basic functions parametrized by $B$. A neural network design with depth K layers is defined in Equation (2):

$$(2)$$
$$Y \approx Y^{NN} = W^T \emptyset_{k-1}(\emptyset_{k-2}(\ldots \emptyset_1(B_1^T \bar{x})))$$

where $\emptyset_k$ and $U_k$ are the element-wise nonlinear function and the weight matrix for the $K^{th}$ layer and W is the output weight matrix as shown in Figure1.

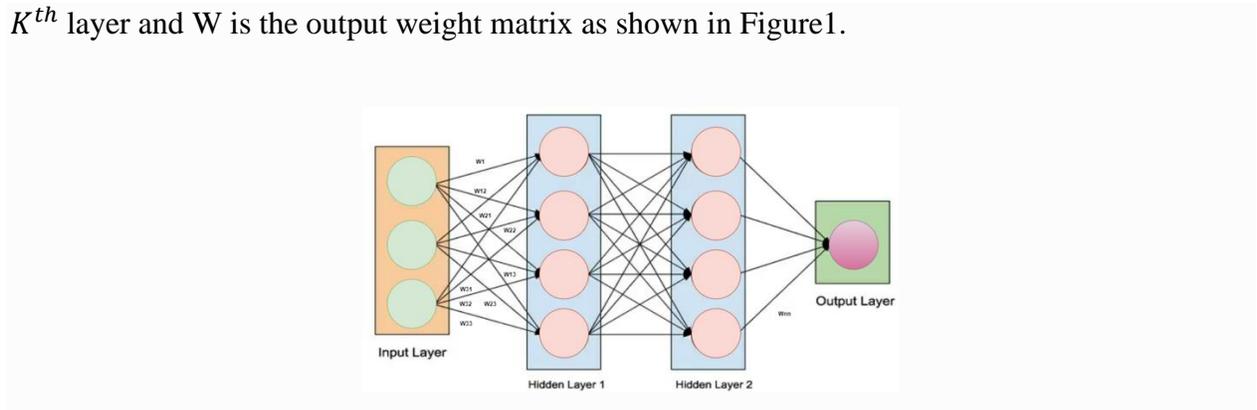

Figure 1 A simple neural network with two hidden layers

Physics-based ML can combine the knowledge we already know, such as physics-based forward and optimization algorithms. The mechanics of training a physics-based network are like training any NNs as shown in Figure 2. It relies on a dataset, an optimizer, and automatic differentiation

[20] to compute gradients. The encoding of information, x, into measurements, y, is given by Equation (3).

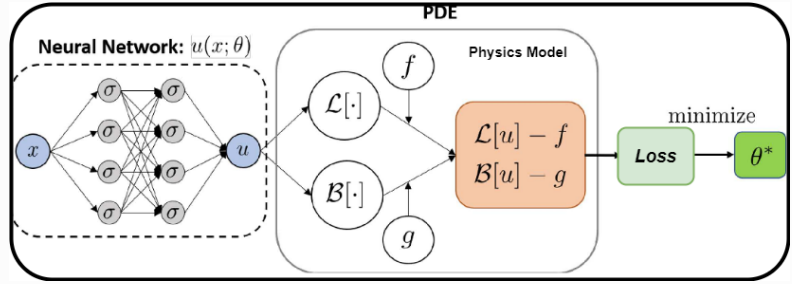

Figure 2. Physics-based ML models are trained to minimize a cost function comprised of equation residual and data over space and time.

$$y = \varpi(x) + \iota \tag{3}$$

where $\varpi$ describes the forward model process that characterizes the formation of measurements and $\iota$ is random system noise. The image reconstruction from a set of measures, i.e., decoding, can be structured using an inverse problem formulation shown in Equation (4).

$$x^* = arg^{min}_x ||\varpi(x) - y||^2 + \daleth(x) \tag{4}$$

where x is the sought information, $||\varpi(x) - y||^2$ is $\mathcal{B}(x; y)$, $\mathcal{B}(.)$ is the data consistency penalty (commonly $\ell_2$ distance between the measurements and the estimated measurements), and $\daleth(.)$ is the signal prior (e.g., sparsity, total variation). This optimization problem often requires a nonlinear and iterative solver. In a nonlinear signal prior, proximal gradient descent can efficiently solve the optimization issue[21]. When numerous constraints are imposed on the picture reconstruction, and the forward model process is linear methods such as the Alternating Direction Method of Multipliers (ADMM) [22] and Half-Quadratic Splitting (HQS)[23] can be efficient solutions. The physics-based network (Figure 3) is created by unrolling N optimization algorithm iterations into network layers. The measurements and initialization are sent into the network, and the output is an estimate of the information after N iterations of the optimizer.

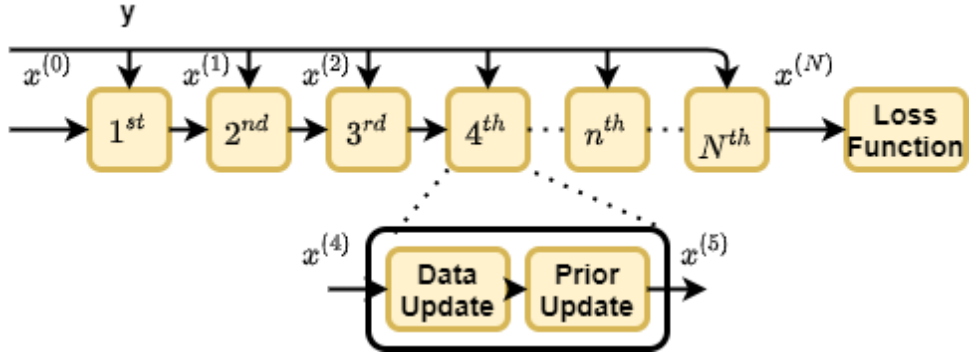

*Figure 3. N unrolled decoder iterations make up an unrolled physics-based network. A data consistency update (e.g., gradient update) and a previous update are included in each layer (e.g., proximal update). The network's inputs are the measurements, y, and initialization, $x^{(0)}$ and the network's output is the final estimate, $x^{(N)}$,. Finally, the final estimate is put into a training loss function.*

## 2. Reduced-order models

Computational Mechanics is a branch of research that needs significant processing power to provide correct results [24]. It nearly always uses a geometric mesh, and the coarseness of the mesh is proportional to the time it takes for a simulation to converge[24]. As a result, it has the potential to grow to such proportions that methods for reducing its order must be developed. These methods aim to create a Reduced-Order Model (ROM) that can effectively replace its heavier counterpart for tasks such as design and optimization, as well as real-time predictions, which all require the model to run many times, which is typically impossible due to a lack of adequate and available computer resources[25]. They capture the behavior of these source models so that civil engineers can quickly study a system's dominant effects using minimal computational resources. ROMs have become popular in civil engineering because engineers face market demands for shorter design cycles that produce higher-quality products and structures. ROMs can be used in civil engineering to simplify various models from full 3D simulations of systems. As a result, civil engineers can use them to optimize structure designs and create more extensive structural simulations

## 3. Proper Orthogonal Decomposition

Partial differential equations (PDEs) are solved using the singular value decomposition (SVD) method. The SVD algorithm applied to PDEs is Proper Orthogonal Decomposition (POD)[16,17].

It's one of the most effective dimensionality reduction techniques for studying complicated spatiotemporal systems[26]. Despite its introduction a few decades ago, POD-based ROM is still state-of-the-art in model order reduction, especially when coupled with Galerkin projection [19]. The dominating spatial subspaces are extracted from a dataset using POD. Put another way, POD calculates the prevailing coherent paths in an infinite space that best describe a system's spatial evolution. Thus, the SVD or eigenvalue decomposition of a snapshot matrix is both strongly connected to POD-ROM. Figure 4 illustrates the evolution of ROM methods.

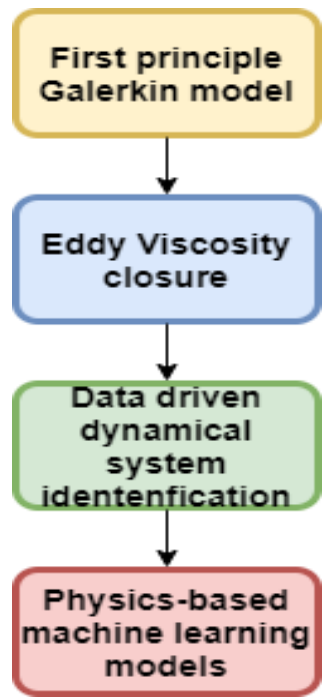

*Figure 4. Evolution of the ROM methods*

### 3.1. Reduced basis through Proper Orthogonal Decomposition

According to [27], a popular technique to create a ROM is to compress it into a smaller area, described by a Reduced Basis set (RB). For the most part, RB techniques follow an offline-online paradigm, with the first being more computationally intensive and the second being quick enough to allow for real-time predictions. The idea is to collect data points from simulation, or any high-fidelity source, called snapshots and stored in an ensemble $\{u^i\}$, and extract the information that has a broader impact on the system's dynamics, the modes, via a reduction method in the offline stage. According to [28, 29] aims at finding a basis function $\varphi(k)$ in a Hilbert space $\mathcal{H}$ possessing the structure of an inner product $(\cdot,\cdot)$ that would optimally represent the field $u$. Supposing that solutions or high-fidelity measures of these solutions are available as $\{u^i\}$, each belonging to the same space $\mathcal{H}$, so that the field can be approximated in Equation (5).

$$u = \sum_{k=1}^{\infty} v^{(k)} \varphi^{(k)} \qquad (5)$$

The Hilbert space for scalar or complex-valued functions is $\mathcal{H} = L^2(\Theta)$, where a space vector x in the domain $\Theta$ is considered, as well as the time variable t, which has an inner product defined by $(f,g) = \int_\Theta f_i g_i^T dx$ (T denotes conjugate transpose). As a result of the summing, variable separation would be possible, as shown in Equation (6).

$$u(x,t) = \sum_{k=1}^{\infty} v^{(k)}(t) \varphi^{(k)}(x) \qquad (6)$$

Considering the mean $<\cdot>$ though as "an average over several separate experiments," e.g., in the case of a function $f$ with $N_r$ realizations $f_i$, $<f> = \frac{1}{N_r}\Sigma_i f_i$, the absolute value $|\cdot|$, and the 2-norm $||\cdot||$ defined as $||f|| = (f,f)^{1/2}$, each normalized optimal basis $\varphi$ is sought after and shown in Equation (7).

$$\max_{\varphi \in \mathcal{H}} \cdot \frac{<|(u,\varphi)|^2>}{||\varphi||^2} \qquad (7)$$

It may be demonstrated to be equal to solving the eigenvalues $\xi$ problem using a condition on variations calculus is shown in Equation (8).

$$\int_\Theta u(x,t) u^*(x',t) \varphi(x') \, dx' = \xi \varphi(x) \qquad (8)$$

The SVD technique converts these continuous expressions to a low-rank approximation in most cases[30]. This method is very comparable to [31] statistical methodology of Principal Component Analysis(PCA), which was evaluated lately [32]. Moreover, this technique may produce a realistic approximation by truncating the sum in Equation.4 to a finite length L, which was originally shown by Sirovich (1987) and represented as shown in Equation (9).

$$u^{POD}(x,t) = \sum_{k=1}^{L} v^{(k)}(t)\varphi^{(k)}(x) \tag{9}$$

Next, the online stage entails retrieving the expansion coefficients and projecting them into our uncompressed, real-life space. Again, the distinction between intrusive and nonintrusive approaches becomes apparent here. The first employs strategies tailored to the problem's formulations, whereas the latter attempts to infer the mapping statistically by treating the snapshots as a dataset.

### 3.2. Intrusive reduced-order methods using the Galerkin procedure.

The Galerkin process is the traditional way to handle the second half of the POD approach to reduced-order modeling, as described and modified [33]. For example, consider the following PDEs, defined by the nonlinear operator $\mathcal{N}$ (Normal distribution) and have the x and t subscripts indicating the associated derivatives shown in Equation (10).

$$u_t = \mathcal{N}_x u \tag{10}$$

The Galerkin technique is used to find each expansion coefficient $v^{(k)}$ from the L-truncated sum in Equation 7. A system of solvable equations is generated by reinjecting the estimated $u^{POD}$ within Equation 8 and multiplying by the L POD modes $\varphi$, known as Galerkin projection. For the $p^{th}$ expansion coefficient, with $\mathcal{R}$ the nonlinear residuals as shown in Equation (11).

$$v_t^{(p)} = \sum_{k=1}^{L} \varphi^{(p)} \mathcal{N}_x u^{POD} \approx \mathcal{R}^{(p)} u^{POD} \tag{11}$$

In [34], this POD-Galerkin method was used to Shallow Water equations such as dam failure and flood forecasts. However, since $\mathcal{R}$ is a generic nonlinear operator, as indicated in these and many other publications, it is unclear how to achieve any speedup in the offline stage, i.e., solving Equation 7, Unless $\mathcal{R}$ is used to make certain approximations. In addition, the reduced basis is parameter-dependent for parameter-dependent issues requiring several simulations, as is the case with uncertainty quantification problems. Therefore, the usage of many RB may be necessary, and finding a way to combine these bases to find an accurate solution is a difficult task[35, 36].

### 3.3. Nonintrusive reduced-order methods using Polynomial Chaos Expansion

A modeling method must be used to make sense of this snapshot collection and create a surrogate model to retrieve the projection coefficients accurately. While traditional and easy approaches like polynomial interpolation appear promising for this job, as pointed out in [37], they struggle to produce useful results with few samples. A different take has been explored within the Polynomial Chaos Expansion (PCE) realm, proposed in [38]. Using Hermite polynomials, and more precisely, a set of multivariate orthonormal polynomials Φ, Wiener's Chaos theory allows for modeling the outputs as a stochastic process. Considering the previous expansion coefficients $v(k)$ $(t)$ as a stochastic process of the variable $t$, the PCE is shown in Equation (12).

$$v^{(k)}(t) = \sum_{\alpha \epsilon C^L} C_\alpha^{(k)} \Phi_\alpha(t) \tag{12}$$

with $\alpha$ identifying polynomials following the right criteria in a set $C^L$ [39]. However, stability issues may arise, and a new approach using the B-Splines Bézier Elements based Method (BSBEM) to address this has been developed in [36]. Unfortunately, while it has shown excellent results, this approach can also suffer from the curse of dimensionality, a term coined half a century ago [40], that still has significant repercussions nowadays, as shown in [41]. In basic terms, it means that many well-intentioned techniques work effectively in narrow domains but have unexpected and unworkable consequences when applied to larger settings.

### 3.4. Data driven methods ROMs

Data driven methods is aiding in the design of ROMs for greater accuracy and cheaper processing costs in various ways. One way is to create an data driven based surrogate model for full-order models[42], where the data driven model may be thought of as a ROM. Two further approaches are to develop an data driven model to replicate the dimensionality reduction mapping from a full-order model to a reduced-order model[43] or to generate an data driven-based surrogate model of an already built ROM using another dimensionality reduction technique[44]. Data driven and ROMs can also be linked by using the data driven model to learn. The residual between observational data and a ROM[45]. The Integration of Physics-Based Modeling and data driven models can significantly increase ROMs' capabilities due to their typically rapid forward execution speed and ability to use data to simulate high-dimensional phenomena. The evolution of ROM is seen in Table 1.

Table 1 The evolution of ROM

| Year | Reference | Key Contributions |
| --- | --- | --- |
| 1915 | [46] | Galerkin method for solving (initial) boundary value problems |
| 1962 | [47] | Low dimensional modeling (with 7 modes) |
| 1963 | [48] | Low dimensional modeling (with 3 modes) |
| 1967 | [49] | Proper orthogonal decomposition (POD) |
| 1987 | [50] | Method of snapshots |
| 1988 | [51] | First POD model: Dynamics of coherent structures and global eddy viscosity modeling |
| 1994 | [52] | Linear modal eddy viscosity closure |
| 1995 | [53] | Gappy POD |
| 2000 | [54] | Galerkin ROM for optimal ow control problems |
| 2001 | [55] | Numerical analysis of Galerkin ROM for parabolic problems |
| 2002 | [56] | Balanced truncation with POD |
| 2003 | [57] | Guidelines for modeling unresolved modes in POD {Galerkin models |
| 2004 | [58] | Spectral viscosity closure for POD models |
| 2004 | [59] | Empirical interpolation method (EIM) |
| 2005 | [60] | Spectral decomposition of the Koopman operator |
| 2007 | [61] | Reduced basis approximation. |
| 2007 | [62] | ROM for four-dimensional variational data assimilation |

| 2008 | [63] | Interpolation method based on the Grassmann manifold approach. |
| 2008 | [64] | Missing point estimation |
| 2009 | [65] | Spectral analysis of nonlinear ows |
| 2010 | [66] | A purely nonintrusive perspective: Dynamic mode decomposition (DMD) |
| 2010 | [67] | Discrete empirical interpolation method (DEIM) |
| 2013 | [68] | The Gauss{Newton with approximated tensors (GNAT) method |
| 2013 | [69] | Proof of global boundedness of nonlinear eddy viscosity closures |
| 2014 | [70] | K-scaled eddy viscosity concept |
| 2015 | [71] | Stabilization of POD Galerkin approximations |
| 2015 | [72] | On bounded solutions of Galerkin models |
| 2016 | [73] | Data-driven operator inference nonintrusive ROMs |
| 2016 | [74] | Spectral POD |
| 2018 | [75] | On the relationship between spectral POD, DMD, and resolvent analysis |
| 2018 | [76] | Shifted/transported snapshot POD. |
| 2018 | [77] | Feature-based manifold modeling |
| 2019 | [78] | Multi-scale proper orthogonal decomposition |
| 2021 | [79] | Cluster-based network models |
| 2021 | [80] | The Potential of Machine Learning to Enchance Computational Fluid Dynamics |
| 2021 | [81] | Towards extraction of orthogonal and parsimonious non-linear modes from turbulent flow |

**4. Development Physics-based machine learning**

Although NNs has been around for a long time, dating back to the [82] perceptron model, they had to wait for the concepts of backpropagation and automatic differentiation, coined [83] and [84], respectively, to have a computationally practical way of training their multilayer, less trivial counterparts. Other types of NNs, such as Recurrent Neural Networks (RNNs) [85] and Long-Short-Term Memory [86] networks, became popular, allowing for advances in sequencing data. While the universal approximation power of DNNs in the context of DL had been predicted for a long time [87], the community had to wait until the early 2010s to finally have both the computational power and practical tools to train these large networks, thanks to for new developments like [88] to mention a few, it rapidly led to advances in making sense of and building upon vast volumes of data. Physics rules are traditionally represented as well-defined PDEs with

Boundary Condition (BC)/Initial Condition (IC) acting as constraints. For example, [ddd-88] developed novel ways in PDEs discovery using just data-driven methodologies [89] and anticipated that this new discipline of DL in dynamic systems like Computational Fluid Dynamics (CFD) would take off (2017)[90]. Its versatility enables various applications, such as missing CFD data recovery [91] or aerodynamic design optimization [39]. The high expense of a fine mesh was solved by using an ML technique to analyze mistakes and adjust amounts in a coarser setting [92]. [93] presented a new numerical scheme, the Volume of Fluid-Machine Learning (VOF-ML) method used in bi-material situations. In addition, research published older research of existing ML algorithms applied to environmental sciences, especially hydrology [94]. Nonetheless, having scant and noisy data at our disposal is typical in engineering, but intuitions or expert knowledge about the underlying physics. It encouraged researchers to consider how to combine the requirement for data in these approaches with system expertise, such as governing equations, first detailed in [95, 96], then extended to NNs in [96] with applications on Computational Fluid Dynamics, as well as in vibrations [97]. A few of these approaches will be explained in detail in below Sections.

### 4.1. Physics-Informed Machine Learning

Despite significant progress in simulating multiphysics problems using numerical discretization of PDEs, it is still impossible to seamlessly incorporate noisy data into existing algorithms, mesh generation is still difficult, and high-dimensional problems governed by parameterized PDEs are unsolvable. Furthermore, tackling inverse issues involving hidden physics is frequently prohibitively costly and necessitates several formulations and complex computer codes. ML has emerged as a viable option, and however, DNNs training needs large amounts of data, which is not always accessible for scientific issues. Instead, additional information acquired by enforcing physical norms may be used to train such networks. This type of physics-informed learning combines (noisy) data with mathematical models, then implemented using NNs or other kernel-based regression networks. Furthermore, customized network designs that automatically meet specific physical invariants for increased accuracy, training speed, and generalization may be conceivable.

### 4.2. Encoding physics in Gaussian Processes

A Gaussian process (GP) is a set of random variables with a Gaussian distribution for a finite number. The mean ($\mathbb{E}$) and covariance functions of a GP define it entirely. The mean function $m(x)$ and the covariance function $k(x, x')$ of a real process $f(x)$ are defined by Equations (12) and (13), respectively.

$$m(x) = \mathbb{E}[f(x)] \tag{12}$$

$$k(x, x') = \mathbb{E}[f(x) - m(x))(f(x') - m(x'))] \tag{13}$$

An Equation defines the Gaussian process. (14).

$$f(x) \sim GP(m(x), k(x, x') \tag{14}$$

"The Gaussian probability distribution is an extension of the Gaussian process. *GP* are nonparametric function estimators with a lot of power. However, when the training data are insufficient to reflect the complexity of the system (generating the data) or the test points are far distant from the training instances (extrapolation), *GP* might perform poorly as a data-driven method. On the other hand, physics information is stated as differential equations and is utilized to create physical models for various research and engineering applications [98]. These models are designed to represent the system's underlying mechanism (i.e., physical processes) and are not constrained by data availability: they can generate accurate predictions even without training data, [99]. For example Bayesian Optimization based on Gaussian process regression is applied to different CFD problems which can be of practical relevance like shape optimization[100]. It has a mean (here 0) and a covariance function $k$, for instance, the Square Exponential. It could be thought of as a very long vector containing every function value $y_i = f(x_i)$ defined as, with $f'$ representing the test outputs of $x'$, not yet observed, demonstrated by Equations (15) and (16).

$$f(x) \sim GP(0, k(x, x'; \theta)) \tag{15}$$

$$\begin{bmatrix} f \\ f' \end{bmatrix} \sim \mathcal{N}\left(0, \begin{bmatrix} k(x, x; \theta) & k(x, x'; \theta) \\ k(x', x; \theta) & k(x', x'; \theta) \end{bmatrix}\right) \tag{16}$$

And the covariance can be, for instance, Gaussian, shown in Equation (17).

$$k\left(x, x'; \begin{bmatrix} \alpha \\ \beta \end{bmatrix}\right) := \alpha^2 exp\left(-\frac{1}{2}\sum_{d=1}^{n} \frac{(x_d - x'_d)^2}{\beta_d^2}\right) \quad (17)$$

## A) Example problem Setup for linear PDEs

Let's now consider time-dependent linear PDEs, as presented in [95]. First, forward Euler gets the result using the simplest temporal discretization method shown in Equation (18).

$$u_t = \mathcal{L}_x u, \quad x \in \Omega, \quad t \in [0, T]$$

$$u^n = u^{n-1} + \Delta t \mathcal{L}_x u^{n-1} \quad (18)$$

and then placing a GP prior shown in Equation (19).

$$u^n(x) \sim GP(0, k_{u,u}^{n-1,n-1}(x, x', \theta)) \quad (19)$$

As a result, the Euler rule is captured in the following multi-output GP, Equation (20). Table 2 gives a pseudo-code of the steps in PINNs involved.

$$\begin{bmatrix} u^n \\ u^{n-1} \end{bmatrix} \sim GP\left(0, \begin{bmatrix} k_{u,u}^{n,n} & \cdots & k_{u,u}^{n,n-1} \\ \vdots & \ddots & \vdots \\ & \cdots & k_{u,u}^{n-1,n-1} \end{bmatrix}\right) \quad (20)$$

Table 2. Implementing a PINNs is straightforward with modern tools

| | |
|---|---|
| 1 | Train hyperparameters $\theta$ with initial $\{x^0, u^0\}$ and boundary $\{x_b^1, u_b^1\}$ data. |
| 2 | Predict artificial data $\{x_b^1, u_b^1\}$ of the next time-step from the posterior. 24 |
| 3 | Train new hyperparameters for the time-step 2, using these artificial data $\{x^1, u^1\}$ and the boundary data $\{x_b^2, u_b^2\}$. |
| 4 | Predict new artificial data $\{x^2, u^2\}$ with these new hyperparameters. |
| 5 | Repeat 3. and 4. until the final time-step. |

**B) Example problem setup for nonlinearity PDEs**

What if $\mathcal{L}_x$ is nonlinear? For example, Burgers' equation is shown in Equation (21).

$$u_t + uu_x = vu_{xx} \text{ with } \mathcal{L}_x := vu_{xx} - uu_{xx} \tag{21}$$

Applying Backward Euler gives the following Equation (22).

$$u^n = u^{n-1} - \Delta t u^n \frac{d}{dx} u^n + v\Delta t \frac{d^2}{dx^2} u^n \tag{22}$$

Assuming un as a GP will not work here since the nonlinear term $u^n \frac{d}{dx} u^n$ will not result in a GP. The idea is to utilize the preceding step's posterior mean, $\mu^{n-1}$ as shown in Equation (23).

$$u^n = u^{n-1} - \Delta t \mu^n \frac{d}{dx} u^n + v\Delta t \frac{d^2}{dx^2} u^n \tag{23}$$

The cubic scaling of computational power with the number of training points due to matrix inversion when forecasting and the necessity to address nonlinear equations on a case-by-case basis are limitations of this technique. This has encouraged scientists to investigate DNNs with built-in nonlinearities.

### 4.3. Physics-Informed Neural Networks

Modeling physical processes described by PDEs has improved thanks to PINNs significantly. The behavior of complicated physical systems is learnt by minimizing the residual of the underlying PDEs by optimizing network settings. PINNs use basic designs to understand the behavior of complicated physical systems by adjusting network settings to reduce the residual of the underlying PDEs. As [96] presented, let's consider generic, parametrized nonlinear PDEs shown in Equation (24).

$$u_t + \mathcal{N}_x^\gamma u = 0, x \in \Omega, t \in [0, T] \tag{24}$$

Whether we aim to solve it or identify the parameters $\gamma$, the idea of the paper is the same: approximating $u(t, x)$ with DNNs, therefore defining the resulting PINNs network $f(t, x)$ is shown in Equation (25).

$$f := u_t + \mathcal{N}_x^\gamma u \tag{25}$$

Now, we'll derive this network using automated differentiation, a chain-rule-based approach famously utilized in typical DL settings, eliminating the requirement for numerical or symbolic differentiation in our situation. Burgers' equation, 1D with Dirichlet IC/BC, is used as a test case as shown in Equations (26), (27), and (28).

$$u_t + uu_x - (0.01/\pi)uu_x = 0, x\epsilon[-1,1], t\epsilon[0,1] \tag{26}$$

$$u(0, x) = -sin(\pi x) \tag{27}$$

$$u(t, -1) = u(t, 1) = 0 \tag{28}$$

From this can define $f(t, x)$, the PINNs is shown in Equation (29).

$$f := u_t + uu_x - (0.01/\pi)uu_x \tag{29}$$

The shared parameters are learned minimizing a custom version of the commonly used Mean Squared Error loss, with $\{t_u^i, x_u^i, u^i\}_{i=1}^{N_u}$ and $\{t_u^i, x_f^i\}_{i=1}^{N_f}$ respectively the IC/BC on $u(t, x)$ and collocations points for $f(t, x)$ is shown in Equation (30).

$$MSE = \frac{1}{N_U}\sum_{i=1}^{N_u} |u(t_u^i, x_u^i) - u^i|^2 + \frac{1}{N_f}\sum_{i=1}^{N_f} |f(t_f^i, x_f^i)|^2 \tag{30}$$

Figure 5 shows the overview of PINNs, which were created using [96]. Table 3 offers a pseudo-code.

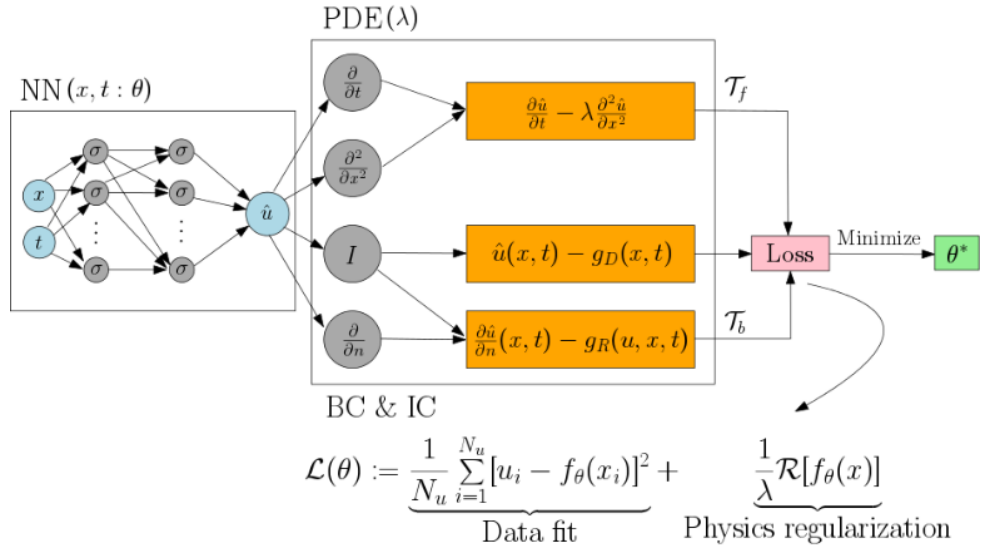

Figure 5. Overview of the PINNs

| | |
|---|---|
| Table 3. Implementing a PINNs is straightforward with modern tools | |
| 1 | Function u(t,x): |
| 2 | $\hat{u} = NN([x, t])$ |
| 3 | Return $\hat{u}$ |
| 4 | |
| 5 | Function f(t,x): |
| 6 | $\hat{u} = u([x, t])$ |
| 7 | $\hat{u}_t = tf.gradients(\hat{u}, t)$ |
| 8 | $\hat{u}_x = tf.gradients(\hat{u}, x)$ |
| 9 | $\hat{u}_{xx} = tf.gradients(\hat{u}_x, t)$ |
| 10 | $\hat{f} = u_t + uu_x - (0.01/\pi)\hat{u}u_x$ |
| 11 | Return $\hat{f}$ |

The same authors have performed further work, applying the framework to different fields, including DL of vortex-induced vibrations [97].

**4.4. The advantages and disadvantages of physics-based ML**

The ability of NNs to approximate solutions to PDEs has been a fascinating field of research. The prediction of dynamics over very long durations that surpass the training horizon over which the network was tuned to represent the solution remains a significant issue. Due to their desirable features, current ML methods, particularly DNNs, have significantly succeeded across computational science areas. First, a sequence of universal approximation theorems [94-96] shows that NNs can approximate any Borel measurable function on a compact set with arbitrary precision given enough hidden neurons. Given enough samples and processing resources, this strong character allows the NNs to approximate any well-defined function.

Furthermore, [101] and more recent studies [102, 103] estimate the convergence rate of approximation error on an NNs with respect to its depth and width, which subsequently allow the NNs to be used in scenarios with high requirements accuracy. The PINNs is applied for solving the Navier-Stokes equation for laminar flows by solving the Falkner-Skan boundary layer results shows the excellent applicability of PINNs for laminar flows with strong pressure gradient [104]. Secondly, the development of differentiable programming and automatic differentiation enables efficient and accurate calculation of gradients of NNs functions with respect to inputs and parameters. These backpropagation algorithms allow the NNs to be efficiently optimized for specified objectives. The following characteristics of NNs have sparked interest in using them to solve PDEs. One general classification of such methods is two classes: The first focuses on directly learning the PDEs operator[105, 106]. For example, in the Deep Operator Network (DeepONet), the input function can be the IC/BC and parameters of the equation mapped to the output, the PDEs solution at the target spatio-temporal coordinates. In this approach, the NNs are trained using independent simulations and must span the space of interest. Therefore, NNs training is predicated on many solutions that may be computationally expensive to obtain. Still, once trained, the network evaluation is computationally efficient[107, 108]. The second class of methods adopts the NNs as a basis function to represent a single solution. The inputs to the network are generally the spatio-temporal coordinates of the PDEs, and the outputs are the solution values at the given input coordinates.

The NNs are trained by minimizing the PDEs residuals and the mismatch in the IC/BC. Such approach dates to[109], where NNs were used to solve the Poisson equation. In later studies [110, 111], the BC was imposed exactly by multiplying the NNs with certain polynomials. In [112], the PDEs are enforced by minimizing energy functionals instead of equation residuals, different from most existing methods. In [96], PINNs for forward and inverse (data assimilation) problems of time-dependent PDEs are developed. To assess all the derivatives in the differential equations and the gradients in the optimization method, PINNs use automated differentiation. Gradients in PINNs are effectively evaluated because automated differentiation consists of analytical derivatives of the activation functions frequently applied in a chain rule. The time dependent PDEs are realized by minimizing the residuals at selected points in the whole spatiotemporal domain. The cost function has another penalty term on the IC/BC if the PDEs problem is forward and a penalty term on observations for inverse data assimilation problems. However, when the underlying PDE solutions contain high-frequencies or multi-scale features, PINNs with fully connected architectures frequently fail to accomplish stable training and provide correct predictions[113, 114] [115]. Recently ascribed this pathological behavior to multi-scale interactions between various components in the PINNs loss function, which eventually lead to stiffness in the gradient flow dynamics, imposing severe stability requirements on the learning rate [116]. Table 4. gives Summary of requirements and possible advantages and disadvantages of physics-based ML.

Table 4. Summary of requirements and possible advantages and disadvantages of physics-based ML.

| Physics-based ML | Requirements | Advantages | Disadvantages |
|---|---|---|---|
| Data | Quality data | Required small amount data compared to non-physics-based ML | Hard to get quality data |
| Cost function | Establish physical relation using PDEs | Physical consistency, improved generalizations, and accuracy | Complex physics PDEs |
| Initialization | Synthetic data from physics models | Reduced observations required, Improved accuracy | Fixed initial state is the resulting exploration challenge |
| Run time | High performance device | Very fast | - |

| Architecture | Based on the complex of task | Intermediate physical variables/processes, Informed prior distributions, Easy to implement using existing packages Such as PINNs, DeepONet | - |

## 5. Application of Physics-based ML to Civil Engineering

There have been applications of physics-based ML models within the field of civil engineering. Figure 6 provides a bar chart that shows the number of ML papers published from 2014 to early 2021, indicating overall growth in papers. This growth seems to follow the rising interest in physics-based ML, which started in 2019, introducing Physics-Informed Neural Networks (PINNs)[96].

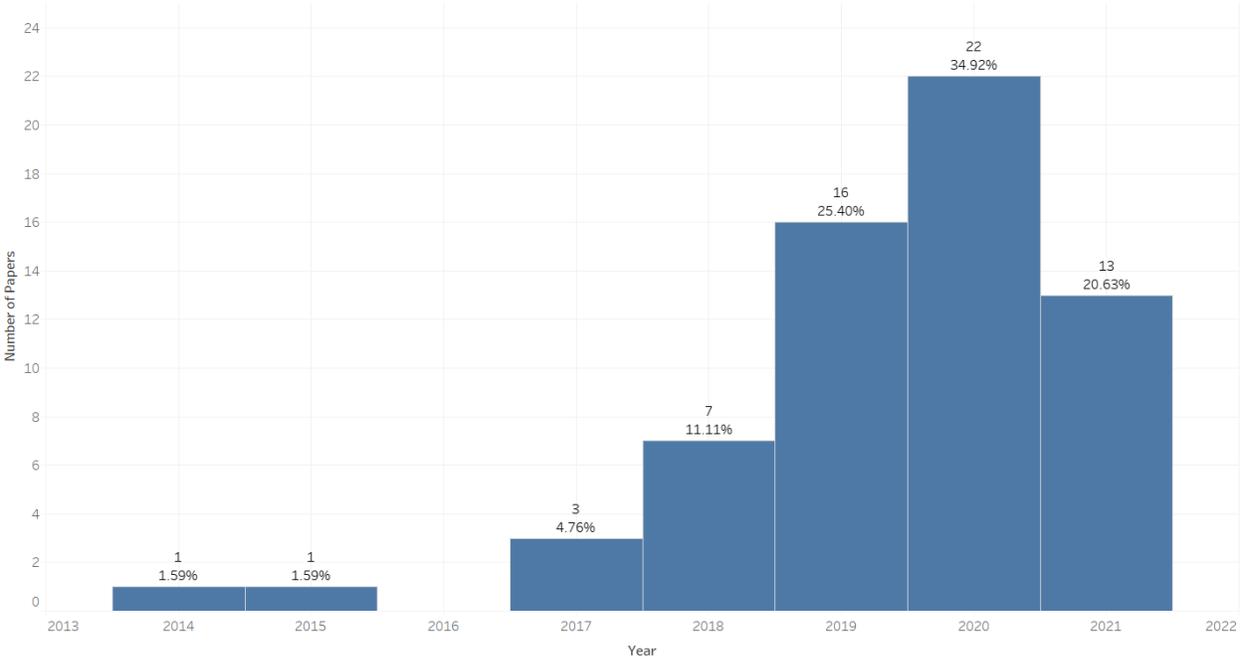

*Figure 6. Papers published from 2014 to early 2021*

Sensor and signal data are mainly used when applying physics-based ML in civil engineering. In contrast, other data sources are employed only based on the requirements. Therefore, researchers must select and reconstruct the algorithms and network structures to solve different civil engineering problems. ML models may learn physics due to their capacity to learn from experience: The ML model can learn how a physical system acts and generate accurate predictions given enough instances of how it behaves. As a result, it may be used in various engineering applications such as damage detection, vibration identification, 3D reconstruction, anomaly data detection, etc.

According to the data types noted in the collected literature, the three main applications of physics-based ML methods in civil engineering are
- 3D Building Information Modelling (BIM)
- Structural health monitoring system
- Structural design and analysis

**5.1 Building Information Modeling (BIM)**

BIM has been highlighted as a new and revolutionary technology for improving the building industry's performance. The BIM tools commonly used in civil engineering are Autodesk's AutoCAD Civil 3D® and Revit Structure ®. These tools optimize and validate projects before they are built and model how infrastructure operates in a 3D real-world setting. To complement 3D BIM, physics-based ML will be a useful tool for problem-solving in geotechnical engineering[117] example shown in Figure 7. The steps carried out in physics-based ML method for 3D BIM were modeling the 3D digital terrain model from point cloud; creating the horizontal alignment, vertical profiles, and editing cross-sections; modeling the jacked tunnel; creating the roundabout; generating the 3D parametric model of the complete road and visualizing the infrastructure in the real-world context governed by PDEs. Table 4 provides a systematic organization and taxonomy of the application-centric objectives and methods of existing physics-based ML for BIM applications.

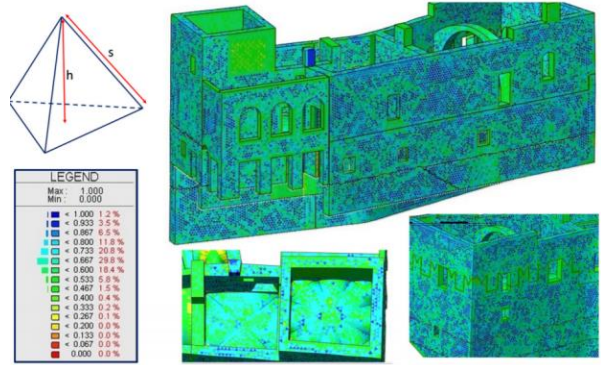

Figure 7 Cloud-to-BIM-to-FEM: Structural simulation with accurate historic BIM from laser scans [118]

*Table 4. Table of literature classified by existing physics-based ML for BIM applications.*

| Paper | Year | Application |
|---|---|---|
| Efficient intensity measures and machine learning classification algorithms for collapse prediction informed by physics-based ground motion simulations[119] | 2020 | CFD, BIM |
| Enhancing predictive skills in a physically consistent way: Physics Informed Machine Learning for Hydrological Processes[120] | 2021 | CFD, BIM |
| Physics-Informed Autoencoders for Lyapunov-stable Fluid Flow Prediction[121] | 2019 | CFD, Turbulence modeling, BIM |
| Predictions of turbulent shear flows using deep neural network[122] | 2019 | CFD, Turbulence modeling |
| Convolutional-network modles to predict wall-bounded turbulence from wall quantities[123] | 2021 | CFD, Turbulence modeling |
| From coarse wall measurements to turbulent velocity fields through deep learning[124] | 2021 | CFD, Trubulence modeling |
| A Data-Driven and Physics-Based Approach to Exploring Interdependency of Interconnected Infrastructure[125] | 2019 | CFD, BIM |
| Physics Guided Machine Learning Methods for Hydrology[126] | 2020 | CFD, BIM |

| Modeling the dynamics of PDE systems with Physics-Constrained Deep Auto-Regressive Network[127] | 2019 | CFD, BIM |
|---|---|---|
| A domain decomposition nonintrusive reduced order model for turbulent flows[128] | 2019 | CFD, BIM |

## 5.2 Structural Health Monitoring

Structural Health Monitoring (SHM) in the civil engineering industry faces unique challenges. These challenges result in part from the dynamic work environments of construction. As a result of its better capacity to detect damage and defects in civil engineering structures, physics-based ML approaches in SHM gained much attention in recent years. Physics-based ML methods establish a high-fidelity physical model of the structure, usually by finite element analysis, and then establish a comparison metric between the model and the measured data from the real structure example shown in Figure 8. In most cases, a vibration-based model updating approach has been chosen, where the vibration or modal data is adopted as the basis for the updating process [6].

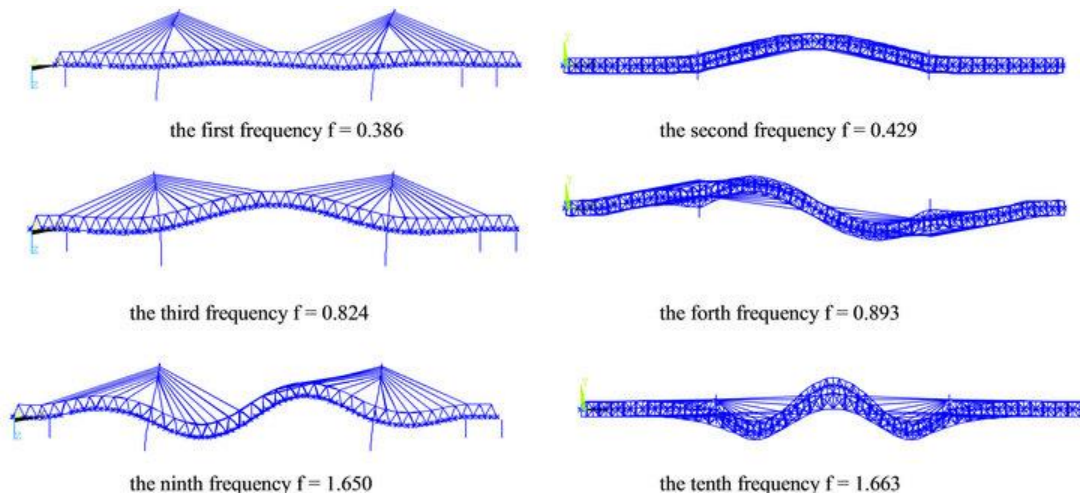

*Figure 8. Vibration Analysis of Vehicle-Bridge System Based on Multi-Body Dynamics using physics-based ML model [114]*

The experimental modal properties of civil engineering structures (say the natural frequencies, vibration modes, and frequency response functions) may be determined by using any of the available system identification methods, such as experimental modal analysis (EMA) or operational modal analysis (OMA). Table 5 provides a systematic organization and taxonomy of the application-centric objectives and methods of existing physics-based ML for SHM applications.

Table 5. Table of literature classified by existing physics-based ML for SHM applications.

| Paper | Year | Application |
| --- | --- | --- |
| Probabilistic physics-guided machine learning for fatigue data analysis[129] | 2020 | SHM |
| Finite element–based machine-learning approach to detect damage in bridges under operational and environmental variations[130] | 2019 | SHM, CFD |
| A hybrid physics-assisted machine-learning-based damage detection using Lamb wave[131] | 2021 | SHM |
| Data-Driven and Model-Based Methods with Physics-Guided Machine Learning for Damage Identification[132] | 2020 | SHM |
| Deep UQ: Learning deep neural network surrogate models for high dimensional uncertainty quantification.[133] | 2018 | Uncertainty quantification, Turbulence modeling, SHM |

5.3 Structural design and analysis

Structures designed with the internal force flow in their members can save a lot of money on materials and labor, but it's difficult and time-consuming. However, the physics-based ML methods have allowed geometry-based structural design methods to reemerge, particularly in three dimensions. The complex geometric diagrams of forces can now be constructed in milliseconds using the current digital computation, allowing structural designers and architects to explore an unexplored realm of efficient spatial structural forms in 3D. The new design strategy A physics-based ML technique that considers structural performance and construction limitations to speed up topological design example shown in Figure 9. Table 6 provides a systematic organization and taxonomy of the application-centric objectives and methods of existing physics-based ML for structural design and analysis applications.

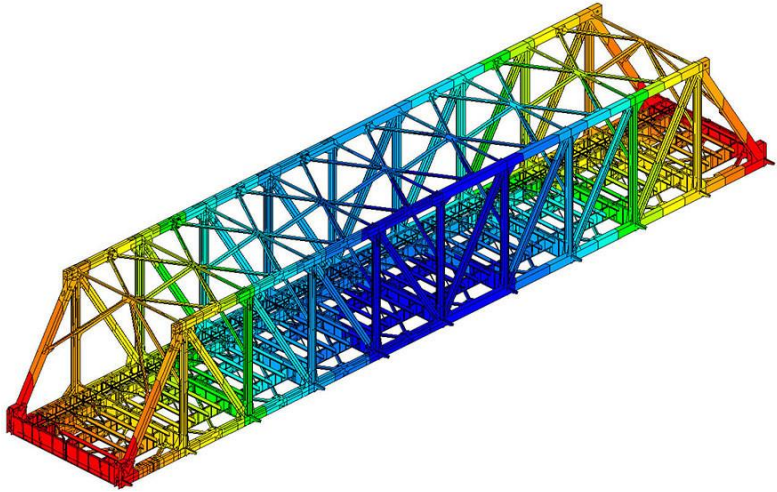

Figure 8. The nonlinear effects of different design variables (i.e., subdivision rules) on the final structural permanence measures, using self-organizing maps for metal bridge using physics-based ML [134]

Table 6. Table of literature classified by existing physics-based ML for structural design and analysis applications.

| Paper | Year | Application |
|---|---|---|
| Machine learning assisted evaluations in structural design and construction[134] | 2020 | Materials science |
| Utilizing physics-based input features within a machine learning model to predict wind speed forecasting error[135] | 2021 | Power system state estimation, Aerodynamics |
| Application of Physics-Based Machine Learning in Combustion Modeling[136] | 2019 | CFD |
| JUNIPR: a framework for unsupervised machine learning in particle physics[137] | 2018 | CFD |
| Machine learning for metal additive manufacturing: predicting temperature and melt pool fluid dynamics using physics-informed neural networks[138] | 2021 | CFD |
| Predicting the dissolution kinetics of silicate glasses by topology-informed machine learning[139] | 2019 | Materials science |
| Machine learning techniques for detecting topological avatars of new physics[140] | 2019 | Materials science |
| Physics-informed machine learning for composition–process–property design: Shape memory alloy demonstration[141] | 2020 | Materials science |
| A novel ozone profile shape retrieval using a full-physics inverse learning machine (FP-ILM)[142] | 2017 | Materials science |
| Machine-learning prediction of thermal transport in porous media with physics-based descriptors[143] | 2020 | Materials science |
| Deep shape from polarization[144] | 2019 | Materials science |
| Model order reduction assisted by deep neural networks(ROM-net)[145] | 2020 | Structural mechanics, Materials science |
| Predicting AC Optimal Power Flows: Combined Deep Learning and Lagrangian Dual Methods[146] | 2019 | Electrical power systems |
| Deep Fluids: A Generative Network for Parameterized Fluid Simulations[147] | 2018 | CFD |
| Multi-Fidelity Physics-Constrained Neural Network and Its Application in Materials Modeling[148] | 2019 | Structural mechanics, Materials science |
| HybridNet: Integrating Model-based and Data-driven Learning to Predict Evolution of Dynamical Systems[149] | 2018 | CFD |
| A composite neural network that learns from multi-fidelity data: Application to function approximation and inverse PDE problems[150] | 2020 | Geosciences |
| PPINN: Parareal physics-informed neural network for time-dependent PDEs[151] | 2020 | CFD, Structural mechanics |
| A deep learning-based approach to reduced-order modeling for turbulent flow control using LSTM neural networks.[43] | 2018 | CFD |
| Physics-induced graph neural network: An application to wind-farm power estimation.[152] | 2019 | CFD |
| Physics-based convolutional neural network for fault diagnosis of rolling element bearings[153] | 2019 | Materials science, Structural mechanics |
| Machine learning closures for model order reduction of thermal fluids[154] | 2018 | Heat transfer |
| Physics-informed machine learning approach for reconstructing Reynolds stress modeling discrepancies based on DNS data[155] | 2016 | Materials science, Structural mechanics |
| A reduced-order model for turbulent flows in the urban environment using machine learning.[44] | 2019 | CFD |
| A Framework for Modeling Flood Depth Using a Hybrid of Hydraulics and Machine Learning[156] | 2020 | CFD |
| Evaluation and machine learning improvement of global hydrological model-based flood simulations.[23] | 2019 | CFD |

| Real-time power system state estimation via deep unrolled neural networks.[157] | 2018 | Power system state estimation |
|---|---|---|
| Symplectic ODE-Net: Learning Hamiltonian Dynamics with Control.[158] | 2019 | CFD |
| Physics-guided Convolutional Neural Network (PhyCNN) for Data-driven Seismic Response Modeling.[159] | 2019 | Structural mechanics, Materials science |

## 6. Future directions

Civil engineering design and construction, which is already a labor-intensive industry, face many challenges, including an aging workforce, increased labor costs, productivity losses, and the lack of onsite workers. All of these constraints affect industry profits. Under these circumstances, physics-based ML will inevitably be utilized to automate some civil engineering and construction processes. Data plays a crucial role in the applications of physics-based ML in civil engineering. Therefore, it is essential to establish a public data set for civil engineering. For example, a similar general-purpose dataset called ImageNet has extensively promoted research in the DL field so that a construction-related dataset could do the same for construction automation. With these kinds of public data sets, researchers can focus more on physics-based ML models.

## 7. Conclusions

As of 2000, ML technology has gradually received more attention in civil engineering and plays an increasingly important role in developing automated technologies. However, the application of even the state-of-the-art black-box ML models has often been met with limited success in civil engineering due to their large data requirements, inability to produce physically consistent results, and their lack of generalizability to out-of-sample scenarios. The main challenges are quality data acquisition and overcoming the impact of the site environment. After thoroughly researching the literature on this topic, this paper suggests that multiple teams could jointly establish an extensive and complete database with the same annotation rules to ease the dilemma of data acquisition. At present, researchers in civil engineering have primarily implemented ML as a tool for feature extraction or detection. We envision that merging ML models and physics principles will play an invaluable role in the future of scientific modeling to address the pressing environmental and physical modeling problems in civil engineering. Future research would be to develop a fully understand physics-based ML and combine them with the specific knowledge domains in civil engineering to develop dedicated physics-based ML models for civil engineering applications.